\documentclass[sigconf]{acmart}

\def\BibTeX{{\rm B\kern-.05em{\sc i\kern-.025em b}\kern-.08emT\kern-.1667em\lower.7ex\hbox{E}\kern-.125emX}}
    
\setcopyright{rightsretained}
\acmConference[COLIEE 2019]{COLIEE 2019 workshop: Competition on Legal
   Information Extraction/Entailment }{June 21, 2019}{Montreal, Quebec}
\acmISBN{}
\acmDOI{}

\begin{document}

%

\title{IITP@COLIEE 2019: Legal Information Retrieval Using BM25 and BERT}

%
 \author{Baban Gain}
  \email{gainbaban@gmail.com}
 \affiliation{%
  \institution{Government College of Engineering and Textile Technology, Berhampore}
  \city{Berhampore}
  \state{West Bengal}
  \postcode{742101}
}

 \author{Dibyanayan Bandyopadhyay}
 \email{dibyanayan@gmail.com}
 \affiliation{%
  \institution{Government College of Engineering and Textile Technology, Berhampore}
  \city{Berhampore}
  \state{West Bengal}
  \postcode{742101}
}
 
 \author{Tanik Saikh}
  \email{1821cs08@iitp.ac.in}
 \affiliation{%
  \institution{Indian Institute of Technology, Patna}
  \city{Bihta, Patna}
  \state{Bihar}
  \postcode{801103}
}

\author{Asif Ekbal}
  \email{asif@iitp.ac.in}
 \affiliation{%
  \institution{Indian Institute of Technology, Patna}
  \city{Bihta, Patna}
  \state{Bihar}
  \postcode{801103}
}





\begin{abstract}
Natural Language Processing (NLP) and Information Retrieval (IR) in the judicial domain is an essential task. With the advent of availability domain-specific data in electronic form and aid of different Artificial intelligence (AI) technologies, automated language processing becomes more comfortable, and hence it becomes feasible for researchers and developers to provide various automated tools to the legal community to reduce human burden. The \textit{Competition on Legal Information Extraction/Entailment (COLIEE-2019)} run in association with the \textit{International Conference on Artificial Intelligence and Law (ICAIL)-2019} has come up with few challenging tasks. The shared defined four sub-tasks (i.e. Task1, Task2, Task3 and Task4), which will be able to provide few automated systems to the judicial system. The paper presents our working note on the experiments carried out as a part of our participation in all the sub-tasks defined in this shared task. We make use of different Information Retrieval(IR) and deep learning based approaches to tackle these problems. We obtain encouraging results in all these four sub-tasks.
\end{abstract}
%
%

%

%

%

\maketitle

\section{Introduction}
Language technology based automated system has a great impact and demand in judicial domain. As for example, it will mitigate the problem of manual checking of large amount of previous files to find relevancy to a current case. This kind of system would be very effective to lawyers which will expedite the process of providing verdict on a particular case.  Previously, it took couple of months and/or even a year to get verdict of a particular case. It is observed that there are many cases where practitioners rely on previous court case files to extract information and try to find relevancy with him/her current solving case. In normal practise they do it process manually. However, computerised processing of such judicial domain texts is taxing as the files are very large in size and remain unstructured. This stipulates lot of research in this domain. The \textit{Competition on Legal Information Extraction and Entailment (COLIEE-2019)} an associated event of \textit{(ICAIL)-2019} has defined four problems related to IR in this domain. Task 1 and Task 2 are involved the case law competition, Task 1 and Task 2 are associated with statute law competition. The data for these two types of tasks are extracted accordingly. We participate in all the four tasks (i.e. Tasks 1, Tasks 2,  Tasks 3 and Tasks 4) defined in this competition. 
All of these four tasks are vary emerging and are of utmost priority. \\ \\
\textit{\textbf{Task 1:}} This task is basically legal case retrieval task. Overall the task is: after reading an new case Q, the system has to extract supporting cases \textit{S1, S2, S3, ....Sn} from the provided case law dataset. This will support the decision for Q.\\
\textit{\textbf{Task 2:}} This legal case entailment task, which involves the detection of a paragraph from the existing cases that entails the decision of a new case.\\
\textit{\textbf{Task 3:}} The task 3 is essentially a legal Question-Answering task. The task could be defined as: reading a legal bar examination question Q and extracting a subset of Japanese Civil Code Articles \textit{S1, S2,..., Sn} from the entire Civil Code which are those appropriate for answering the question such that i.e. \\ 
Entails(S1, S2, ..., Sn , Q) or Entails(S1, S2, ..., Sn , not Q).\\
Overall, the task is given a question Q and the entire Civil Code Articles, the system has to retrieve the set of articles \textit{"S1, S2, ..., Sn"} as the answer of this track. \\ 
\textit{\textbf{Task 4:}} This task involves legal question answering task which associates the identification of an entailment relationship such that \\ 
Entails(S1, S2, ..., Sn , Q) or Entails(S1, S2, ..., Sn , not Q). \\ 
Given a question \textit{Q}, the participating systems have to retrieve relevant articles \textit{S1, S2, ..., Sn} in the first phase, and in the second phase, the systems have to determine if the relevant articles entail "Q" or "not Q". The answer of this track would be binary: i.e. \textit{"YES"("Q") or "NO"("not Q")}.\\
We propose various approaches to combat these problems. We discuss all our approaches in the section \ref{SD}. 
\section{Our Approaches}
The propose approaches for the all four tasks are based on \textit{Doc2Vec}, \textit{BM25}, \textit{tf-idf}, \textit{Bidirectional Encoder Representations from  Transformers (BERT)} \cite{DBLP:journals/corr/abs-1810-04805}. The following subsections will discuss all the approaches one by one.
\subsection{Doc2Vec}
Doc2Vec \cite{DBLP:conf/icml/LeM14} is an unsupervised algorithm that learns fixed-length feature representations from variable-length pieces of texts, such as sentences, paragraphs, and even a whole documents. This algorithm represents each document by a dense vector which is trained to predict words in the document. More precisely, we concatenate on the paragraph vector with several word vectors from a paragraph and predict the following word in the given context. Both word vectors and paragraph vectors are trained by the stochastic gradient descent and back-propagation \cite{rumelhart1988learning}. While paragraph vectors are unique among paragraphs, the word vectors are shared. At prediction time, the paragraph vectors are inferred by fixing the word vectors and training the new paragraph vector until convergence. 
\subsection{BM25} 
BM25 \cite{robertson2009probabilistic} is one of the most successful text-retrieval algorithm. BM25 is a bag-of-words retrieval function that ranks a set of documents based on the query terms appearing in each document, regardless of the inter-relationship between the query terms within a document. Given a query Q, containing keywords {q1,q2,....qn} the BM25 score of a document D is:
\begin{equation}
Score(D,Q) = \sum_{i=1}^{n} IDF(q_i)\frac{f(q_i, D). (k_1+1)}{f(q_i, D)+k_i.(1-b+b.\frac{|D|}{avgdl})}
\end{equation}
where f($q_i$, D) is $q_i$'s term frequency in the document D,
|D| is the length of the document D in words, and \textit{avgdl} is the average document length in the text collection from which documents are drawn.  

\subsection{Term Frequency * Inverse Document Frequency}
\textit{Term Frequency and Inverse Document Frequency (tf*idf)} is a numerical statistic that is intended to reflect how important a word is to a document in a collection of corpus. It is often used as a weighting factor in searches of information retrieval, text mining, and user modeling. The tf*idf value increases proportionally to the number of times a word appears in the document and is offset by the number of documents in the corpus that contain the word, which helps to adjust for the fact that some words appear more frequently in general. The formula that is used to compute the tf*idf for a term t of a document d in a document set is:
\begin{equation}
tf-idf(t, d) = tf(t, d) * idf(t)
\end{equation}
We implement this using python library \cite{pedregosa2011scikit}. 
\subsection{Bidirectional Encoder Representations from Transformers (BERT)}
We make use of a novel language representation model called \textit{Bidirectional Encoder Representations from Transformers (BERT)}. The model is the latest language representation model used in various tasks of NLP. The task 4 defined in this shared task is a classification task. We treat the task as a sentence classification task by considering 't1' and 't2' as two sentences respectively. Using BERT we obtain a vector corresponding to the '[CLS]' token at the last layer. This vector is subsequently used for classification. 
So per 't1' and 't2' we get one vector to be subsequently used in a downstream model (e.g a feed-forward neural net layer) for further classification.
We tried with adding 1 or 2 Dense layers but actually the accuracy obtained was not good to report. The vector obtained was of 768 dimensions and the training set contains 724 examples. Each example contains more dimensions of a vector than the number of examples combined. So the data is prone to be over-fitted. So we resort to use a popular machine learning model known as \textit{eXtreme Gradient Boosting (XGBoost)} \cite{chen2016xgboost} for classification. 
XGBoost gained much popularity recently as a winning solution in several machine learning competitions. It is actually a library which provides gradient boosting framework in several languages. We treat each of the dimension of the vector corresponding to '[CLS]' as a separate feature. So we got 724 training examples each with 768 features. We use a matrix representing this data with dimension [724,768]. This is given to the XGBoost model for classification. 

\section{Data}
The whole competition task is divided into two categories  \textit{viz 1. Case Law Competition (Task 1 and task 2) 2. Statute Law Competition (Task 3 and Task 4)}. The task organizer's released the data for all these four tasks defined. The data is obtained from an existing collection of predominantly Federal Court of Canada case law. In task-1 the training data consists of 285 case queries and each query has 200 candidate notice cases. So each query has 9.87 notice cases on average. The training data of task-2 comprises of 181 queries and each query has an average of 48.59 candidate paragraphs for recognizing entailment relation. In the training data of this task on an average 1.32 paragraphs have an entailment relation with a query. For the task 3 and 4 data are borrowed from Japanese civil law articles (translated to English) and the size of the data for these two tasks are same having 1044 articles.  
  
\section{System Description}\label{SD}
There were four sub-tasks in this competition. We attempt all of those. In the following subsections we discussed about the experimental procedures, models etc. 
Basically, the pre-processing module for task 1 and task 2 is the same, which is described in the following paragraph. 

\textbf{\textit{Pre-processing:}} The pre-processing module comprises of two stages. In the first stage, i.e. for the first model (named as IITPd2v), we perform minimal amount of pre-processing. All the candidate cases and the base cases are extracted and the paragraph numbers are removed. In the second stage i.e. for second model (named as iitpBM25), we remove all the words having length lower than 3 character and all the numerical values from every documents. We also remove all the stop-words using NLTK tool. Then we lemmatize the words using NLTK WordNetLemmatizer \footnote{https://pythonprogramming.net/lemmatizing-nltk-tutorial/}. \\

\subsection{Task1}
The goal of this task is to explore and evaluate case law retrieval technologies that are both effective and reliable. The task investigates the performance of systems that search a set of legal cases that support a previously unseen case description. The input of this system is a query and return noticed cases in the given collection as output. We say a case is 'noticed' with respect to a query iff the case supports the decision of the query case. In this task, the query case does not include a decision, because our goal is to determine how accurately a machine can capture decision-supporting cases for a new case (with no decision). \\
The training data consists of 285 numbers case queries for this task, and each query has 200 numbers candidate noticed cases. In the training data of this task, each case query has an average of 9.87 noticed cases. We propose three models for this task namely  \textit{IITP Doc2Vec (IITPd2v)}, \textit{iitpBM25} and \textit{ iitpDocBM}\\

\subsubsection{Models}
We have three models to tackle this problem, we will discuss all these in detail in the following paragraphs.\\
\textbf{\textit{iitpd2v:}} In this model, we train the data obtained from stage-1 of pre-processing module. For training purpose we make use of gensim Doc2Vec with vector size = 150, window = 10, and with 50 epochs. For every query, we compute the Doc2Vec similarity between the base case and all of its candidates. As the task contains variable number of noticed cases, top 10 candidates whose similarity was greater than 90\% of average score of top two are returned. Total 245 documents are returned for 61 base cases. This model yields the precision, recall, and F-measure as 0.4653, 0.3455, and 0.3965 respectively.\\
\textbf{\textit{iitpBM25:}} From this model, we collect the data yielded by the stage-2 of pre-processing module. We compute BM25 score of all the candidate cases for every base case using gensim library. Then the top 10 candidates whose similarity is greater than 90\% of average score of top two are returned. Thus, the total of 203 documents are returned. This model produces the precision, recall, and F-measure as  0.6256, 0.3848, and 0.4765 respectively.\\
\textbf{\textit{iitpDocBM:}} In this model we combine the scores obtained from the previous two models i.e. from the iitpd2v and from the iitpBM25. The make multiplication of those two scores, so that only documents with high scores on both Doc2Vec and BM25 are ranked. Top 10 candidates whose similarity was greater than 80\% of average score of top two are returned. In this way total of 201 documents are returned. We compute the precision, recall, and F-measure and get the scores as  0.6368, 0.3879, and 0.4821 respectively for this model.
\subsection{Task2} 
In this task, the goal is to predict the decision of a new case by entailment from previous relevant cases. As a simpler version of predicting a decision, a decision of a new case and a noticed case will be given as a query. Then a case law textual entailment system must identify which paragraph in the noticed case entails the decision, by comparing the extracting and the meanings of the query and paragraph. \\
\subsubsection{Model}
We propose three models for this task. The name of the models are \textit{viz. i. iitp2D2v ii. iitpBM25 and iii. iitp2DocBM}
\textbf{\textit{iitp2D2v:}} For iitp2D2v module, we take the data from stage-1 of pre-processing module which we train using gensim Doc2Vec. The specifications for the training is as follows: vector size = 100, window = 5, and epochs = 50. For every query, Doc2Vec similarity between the entailed fragment and all of its candidates are computed and the candidate with highest score are returned. This yields the Precision, Recall, and F-measure of 0.0455, 0.0444, and 0.0449 respectively.\\ 
\textbf{\textit{iitpBM25:}}For iitpBM25, we take the data from stage-2 of the pre-processing module. For every entailed fragment BM25 score of all the candidates are computed by employing gensim library. Then candidate with highest score are returned. This system yields the Precision, Recall, F-measure scores of 0.7045, 0.6889, and 0.6966 respectively. Please note that we got the highest Precision score among all the participants in this task. We obtain the F-score which is the second highest among unique teams.\\ 
\textbf{\textit{iitp2DocBM:}} In iitp2DocBM module, the scores obtained from iitp2D2v and iitpBM25 are multiplied, so that only candidates with high scores on both the methods are ranked. Then the candidate with the highest score is returned. We obtain the Precision, Recall, and F-measure score as 0.6591, 0.6444, and 0.6517 respectively in this task. 

\subsection{Task3} 
This task investigates the performance of the systems that search a static set of civil law articles using previously unseen queries. The goal of this task is to return relevant articles in a collection to a query. We call an article as "Relevant" to a query iff the query sentence can be answered as \textit{Yes/No}, i.e. entailed from the meaning of the article. If combining the meanings of more than one article (e.g., "A", "B", and "C") can be answered a query sentence, then all the articles ("A", "B", and "C") are considered as "Relevant". If a query can be answered by an article "D", and it can also be answered by another article "E" independently, we also consider both of these articles "D" and "E" as "Relevant". This task requires the retrieval of all the articles that are relevant to answering a query. \\
\subsubsection{Model} We propose four models in this task namely \textit{iitpBM25, iitpBM25-L, iitptfidf and iitptfidf-L}. The following points will discuss all the models in detail.\\
\textbf{\textit{iitpBM25:}} In this module, for every query BM25 score of all the articles are computed using gensim library. Then candidates having the highest score are returned. The total 98 results are retrieved among which 54 are correct. Total number of correct entries in the corpus was 121. We obtain the Precision, Recall, and F2-measure of 0.5510, 0.4462, and 0.4639 respectively.\\
\textbf{\textit{iitpBM25-L:}} This model is the modified version of the previous one. We extract the top 100 documents in Descending Order of their score for each query. 109 out of 121 articles in the corpus are retrieved correctly. 
\textbf{\textit{iitptfidf:}} In this model, for every query tf-idf score of all the articles are computed using sklearn TfIdfVectorizer library\footnote{https://scikit-learn.org/stable/modules/generated/sklearn.feature\_extraction.text.TfidfVectorizer.html}. The candidates having the highest scores are returned. Total 98 articles are retrieved among which 49 are correct. Total number of correct entries in the corpus was 121. So the precision, recall, and F2-measure provided by this system are 0.50, 0.4049, and 0.4209 respectively.\\ 
\textbf{\textit{iitptfidf-L:}} This model also the modified version of the previous one. We retrieve top 100 documents for each query. The model retrieves 108 articles out of which 121 articles correct.\\
\subsection{Task4}
\textbf{\textit{Pre-processing:}}  In the Task 4 the pre-processing procedure is as follows. We extract the data from the given .xml files containing tags named as 'pairs', 't1', and 't2'. Under a 'pair' of 't1' and 't2', we have to find their entailment relationship. We use the 'id' attribute under the 'pair' tag as a key to store the corresponding article(t1) and the document(t2). For the training of the model we make use of the training data provided for Task 3 and tested the system using the test data provided in the Task 4. We extract the task-4 data for training in similar ways as of task-3 discussed above. Here along with 't1' and 't2', we also got the 'label' attribute under pair tag which are used as training labels. We gather all 't1'\'s and 't2'\'s along with their labels to form the training dataset. In Task 4 for training purpose we got 12 files containing 724 training examples with 353 instances as \textit{'Yes'} and 371 instances \textit{'No'}.\\
\subsubsection{Model}
Here only one model is offered. The description of the model is as follows.\\
\textit{\textbf{iitpbert:}} In this task we propose a model named as \textit{'iitpbert'}. We make use of BERT-Base-Uncased (Bidirectional Encoder Representation from Transformer)\cite{DBLP:journals/corr/abs-1810-04805} model to combat this problem. We provide all the 't1's and 't2's as inputs to the BERT model. This task is basically a sentence pair classification task. So we obtain the final hidden state (i.e. output of Transformer) corresponding to the first \textit{'[CLS]'} token as the vector representation for the classification tasks. We check that the combined length of t1 and t2  should be less than 512 or not. BERT model can handle the maximum length of 512. We strip the tokens after index 512, if it exceeds 512. 
We obtain a matrix of shape N*M where N is number of training examples and M is the dimension of the output vector of BERT (here M = 768). We try to add dense layers and other machine learning models like SVM and XgBoost on top of that output vectors to train them end to end. This yields an accuracy of around 55\% when validated on a held out train set. We choose to stay with the trained XgBoost classifier and use it to predict labels for the task-4. We submit that file with predicted file for evaluation. This approach produces an accuracy of 59.18\% on the test set for this task.
\section{Results and Discussions}
The task organizer's defined various measures to evaluate the participating systems. The task 1 and task 2 rely on precision, recall and F-measure. In task 3, in addition to precision, recall and F2-measure they also encourage the participant to take Mean Average precision (MAP) into account. Whereas, in task 4 only accuracy is provided as the evaluation metric. The results for the task1, and task-2 in three methods are shown in the Table \ref{resu-task1} and Table \ref{resu-task2} respectively.
\begin{table}[]
\centering
\caption{Results obtained by proposed models in Task-1}
\label{resu-task1}
\begin{tabular}{|l|l|l|l|}
\hline
Models & Precision & Recall & F-measure \\ \hline
iitpd2v & 0.4653 & 0.3455 & 0.3965 \\ \hline
iitpBM25 & 0.6256 & 0.3848 & 0.4765 \\ \hline
iitpDocBM & 0.6368 & 0.3879 & 0.4821 \\ \hline
\end{tabular}
\end{table}
\begin{table}[]
\centering
\caption{Results obtained by proposed models in Task-2}
\label{resu-task2}
\begin{tabular}{|l|l|l|l|}
\hline
Models & Precision & Recall & F-measure \\ \hline
iitpd2v & 0.0455 & 0.0444 & 0.0449 \\ \hline
iitpBM25 & 0.7045 & 0.6889 & 0.6966 \\ \hline
iitpDocBM & 0.6591 & 0.6444 & 0.6517 \\ \hline
\end{tabular}
\end{table}
The results for the task-3 using four models are shown in the Table \ref{resu-task3}  

\begin{table}[]
\centering
\caption{Results obtained by four proposed models in Task-3}
\label{resu-task3}
\begin{tabular}{|l|l|l|l|l|}
\hline
Models & Precision & Recall & F2-measure & MAP \\ \hline
iitpBM25 & 0.4898 & 0.3967 & 0.4142 & X \\ \hline
iitpBM25-L & X & X  & X & 0.5409 \\ \hline
iitptfidf & 0.4388 & 0.3554 & 0.3694 & X \\ \hline
iitptfidf-L & X &X  & X & 0.5056 \\ \hline
\end{tabular}
\end{table}
Please note that here we have calculated Mean Average Precision (MAP) by taking top 100 ranked documents into account. The MAP scores are shown in iitpBM25-L and  iitptfidf-L methods in the 5th column of the Table \ref{resu-task3}.  
In task 4 participants are encouraged to compute the accuracy to evaluate their submitted systems. So in this task we compute the accuracy on from our \textit{Yes/No} prediction. We obtain the accuracy of 59.18\% for the task with our proposed model. 

\section{Conclusion and Future work}
The paper presents our experiment's report performed as a part of our participation in all the four sub-tasks defined in the shared task entitled \textit{COLIEE-2019} organized in \textit{ICAIL 2019}. 
We propose various text retrieval methods (Doc2Vec, BM25, and tf-idf ) for sub-task 1, sub-task 2, and sub-task 3. BERT model is employed for tackling the sub-task 4. We obtain encouraging results in all the four sub-tasks and our ranks in all the four task are in a quite modest position among all the participants in the leaderboard. Our future line of research could be something as follows:
\begin{itemize}
    \item Using some good pre-trained vector embedding with sent2vec could produce better result, so we would link to foster this in future.
    \item We would like to use BERT-large model (another version of BERT) to explore how it is performing.
    \item We would like to do retrospective analysis of the best three models in this competition to get some insights and for better improvement further. 
\end{itemize}

\section{Acknowledgements}
Mr Tanik Saikh acknowledges all the co-authors for their contributing efforts. He is also acknowledging the funding agency to provide the fund for the registration (if anything needs to be done). 

\bibliographystyle{ACM-Reference-Format}
\bibliography{sample-base}

%
\appendix









\end{document}